\documentclass[runningheads]{llncs}
\usepackage{graphicx}
\usepackage{amsmath,amssymb} % define this before the line numbering.
\usepackage{color}\usepackage[width=122mm,left=12mm,paperwidth=146mm,height=193mm,top=12mm,paperheight=217mm]{geometry}
\newcommand\etal{\emph{et al}.\ }
\newcommand\ie{\emph{i.e.}\ }
\newcommand\methodA{\textbf{A}}
\newcommand\methodB{\textbf{B}}
\newcommand\methodC{\textbf{C}}
\begin{document}
% \linenumbers
\pagestyle{headings}
\mainmatter
\title{Simultaneous Detection and Segmentation} % Replace with your title

\titlerunning{Simultaneous Detection and Segmentation}

\authorrunning{Bharath Hariharan \and Pablo Arbel\'{a}ez \and Ross Girshick \and Jitendra Malik}

\author{Bharath Hariharan$^1$ \and Pablo Arbel\'{a}ez$^{1,2}$  \and Ross Girshick$^1$  \and Jitendra Malik$^1$\\
\texttt{\small \{bharath2,arbelaez,rbg,malik\}@eecs.berkeley.edu}}
\institute{$^1$University of California, Berkeley \\ $^2$Universidad de los Andes, Colombia}

\maketitle

\begin{abstract}
We aim to detect all instances of a category in an image and, for each instance, mark the pixels that belong to it. We call this task Simultaneous Detection and Segmentation (SDS). Unlike classical bounding box detection, SDS requires a segmentation and not just a box. Unlike classical semantic segmentation, we require individual object instances. We build on recent work that uses convolutional neural networks to classify category-independent region proposals (R-CNN~\cite{GirshickCVPR14}), introducing a novel architecture tailored for SDS. We then use category-specific, top-down figure-ground predictions to refine our bottom-up proposals. We show a 7 point boost (16\% relative) over our baselines on SDS, a 5 point boost (10\% relative) over state-of-the-art on semantic segmentation, and state-of-the-art performance in object detection. Finally, we provide diagnostic tools that unpack performance and provide directions for future work.

\end{abstract}

\section{Introduction}

Object recognition comes in many flavors, two of the most popular being object detection and semantic segmentation. Starting with face detection, the task in object detection is to mark out bounding boxes around each object of a particular category in an image. In this task, a predicted bounding box is considered a true positive if it overlaps by more than 50\% with a ground truth box, and different algorithms are compared based on their precision and recall. Object detection systems strive to find every instance of the category and estimate the spatial extent of each.  However, the detected objects are very coarsely localized using just bounding boxes.

In contrast, semantic segmentation requires one to assign a category label to all pixels in an image. The MSRC dataset~\cite{ShottonECCV06} was one of the first publicly available benchmarks geared towards this task. Later, the standard metric used to evaluate algorithms in this task converged on pixel IU (intersection over union): for each category, this metric computes the intersection over union of the predicted pixels and ground truth pixels over the entire dataset. This task deals with ``stuff" categories (such as grass, sky, road) and ``thing" categories (such as cow, person, car) interchangeably. For things, this means that there is no notion of object instances. A typical semantic segmentation algorithm might accurately mark out the  dog pixels in the image, but would provide no indication of how many dogs there are, or of the precise spatial extent of any one particular dog. 

These two tasks have continued to this day and were part of the PASCAL VOC challenge~\cite{EveringhamIJCV10}. Although often treated as separate problems, we believe the distinction between them is artificial. For the ``thing" categories, we can think of a unified task: detect all instances of a category in an image and, for each instance, correctly mark the pixels that belong to it. Compared to the bounding boxes output by an object detection system or the pixel-level category labels output by a semantic segmentation system, this task demands a richer, and potentially more useful, output. Our aim in this paper is to improve performance on this task, which we call \textbf{Simultaneous Detection and Segmentation} (SDS).

\begin{figure}
\centering
\includegraphics[width=\textwidth]{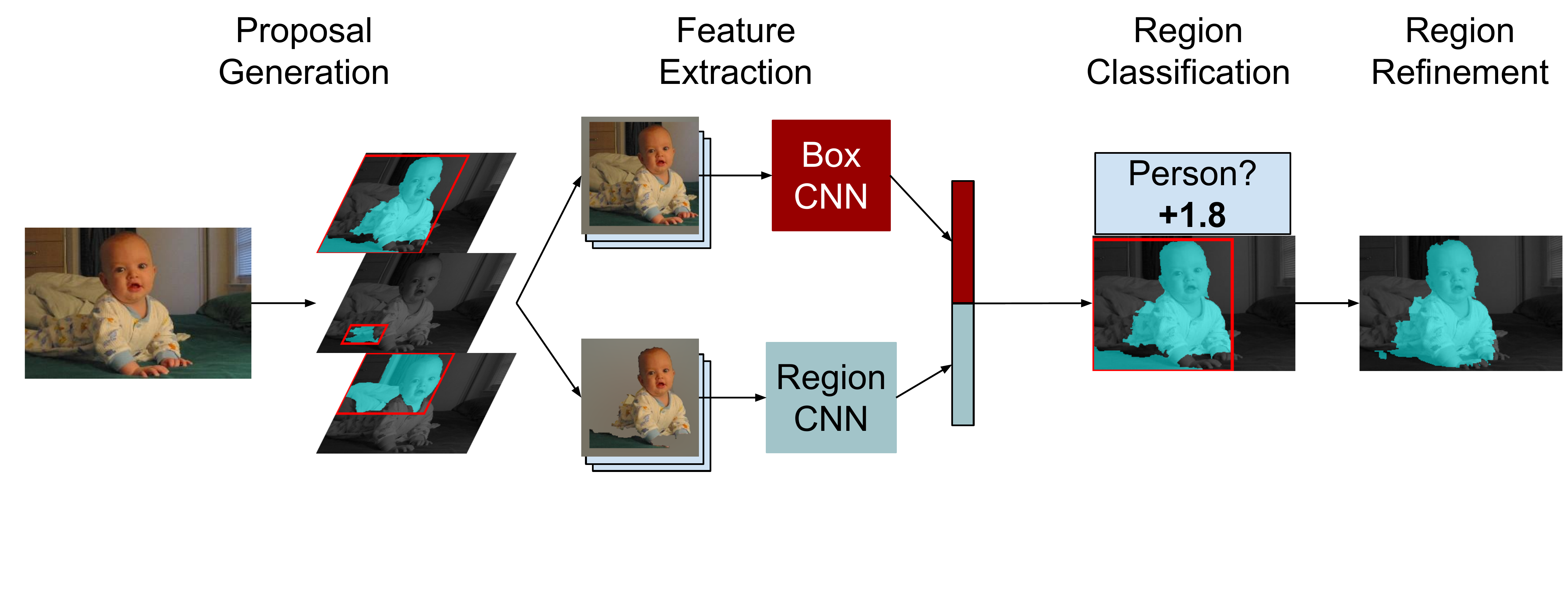}
\caption{Overview of our pipeline. Our algorithm is based on classifying region proposals using features extracted from both the bounding box of the region and the region foreground with a jointly trained CNN. A final refinement step improves segmentation.}
\label{fig:overview}
\end{figure}

The SDS algorithm we propose has the following steps (Figure~\ref{fig:overview}):
\begin{enumerate}
\item \textbf{Proposal generation}: We start with category-independent bottom-up object proposals. Because we are interested in producing segmentations and not just bounding boxes, we need \emph{region} proposals.  We use MCG~\cite{ArbelaezCVPR14} to generate 2000 region candidates per image. We consider each region candidate as a putative object hypothesis.
\item \textbf{Feature extraction}: We use a convolutional neural network to extract features on each region. We extract features from both the bounding box of the region as well as from the region foreground. This follows work by Girshick \etal\cite{GirshickCVPR14} (R-CNN) who achieved competitive semantic segmentation results and dramatically improved the state-of-the-art in object detection by using CNNs to classify region proposals. We consider several ways of training the CNNs. We find that, compared to using the same CNN for both inputs (image windows and region masks), using separate networks where each network is finetuned for its respective role dramatically improves performance. We improve performance further by training both networks jointly, resulting in a feature extractor that is trained end-to-end for the SDS task. 
\item \textbf{Region classification}: We train an SVM on top of the CNN features to assign a score for each category to each candidate. 
\item \textbf{Region refinement}: We do non-maximum suppression (NMS) on the scored candidates. Then we use the features from the CNN to produce category-specific coarse mask predictions to refine the surviving candidates. Combining this mask with the original region candidates provides a further boost.
\end{enumerate}

Since this task is not a standard one, we need to decide on evaluation metrics. The metric we suggest in this paper is an extension to the bounding box detection metric.  It has been proposed earlier~\cite{TigheCVPR14,YangTPAMI12}. Given an image, we expect the algorithm to produce a set of object hypotheses, where each hypothesis comes with a predicted \emph{segmentation}  and a score. A hypothesis is correct if its \emph{segmentation} overlaps with the \emph{segmentation} of a ground truth instance by more than 50\%. As in the classical bounding box task, we penalize duplicates. With this labeling, we compute a precision recall (PR) curve, and the average precision (AP), which is the area under the curve. We call the AP computed in this way AP$^r$, to distinguish it from the traditional bounding box AP, which we call AP$^b$ (the superscripts $r$ and $b$ correspond to region and bounding box respectively). AP$^r$ measures the accuracy of segmentation, and also requires the algorithm to get each instance separately and completely. Our pipeline achieves an AP$^r$ of 49.5\% while at the same time improving AP$^b$ from 51.0\% (R-CNN) to 53.0\%. 

One can argue that the 50\% threshold is itself artificial. For instance if we want to count the number of people in a crowd, we do not need to know their accurate segmentations. On the contrary, in a graphics application that seeks to matte an object into a scene, we might want extremely accurate segmentations. Thus the threshold at which we regard a detection as a true positive depends on the application. In general, we want algorithms that do well under a variety of thresholds. As the threshold varies, the PR curve traces out a PR surface. We can use the volume under this PR surface as a metric. We call this metric AP$^r_{vol}$ and AP$^b_{vol}$ respectively. AP$^r_{vol}$ has the attractive property that an AP$^r_{vol}$ of 1 implies we can perfectly detect and precisely segment all objects. Our pipeline gets an AP$^r_{vol}$ of 41.4\%. We improve AP$^b_{vol}$ from 41.9\% (R-CNN) to 44.2\%.

We also find that our pipeline furthers the state-of-the-art in the classic PASCAL VOC semantic segmentation task, from 47.9\% to 52.6\%. Last but not the least, following work in object detection~\cite{HoiemECCV12},  we also provide a set of diagnostic tools for analyzing common error modes in the SDS task. 
Our algorithm, the benchmark and all diagnostic tools are publicly available at http://www.eecs.berkeley.edu/Research/Projects/CS/vision/shape/sds.

\section{Related work}

For semantic segmentation, several researchers have tried to use activations from off-the-shelf object detectors to guide the segmentation process. Yang \etal\cite{YangTPAMI12} use object detections from the deformable parts model~\cite{FelzenszwalbPAMI10} to segment the image, pasting figure-ground masks and reasoning about their relative depth ordering. Arbel\'{a}ez \etal\cite{ArbelaezCVPR12} use poselet detections~\cite{BourdevECCV10} as features to score region candidates, in addition to appearance-based cues. Ladicky \etal\cite{LadickyECCV10} use object detections as higher order potentials in a CRF-based segmentation system: all pixels in the foreground of a detected object are encouraged to share the category label of the detection. In addition, their system is allowed to switch off these potentials by assigning a true/false label to each detection. This system was extended by Boix \etal\cite{BoixIJCV12} who added a global, image-level node in the CRF to reason about the categories present in the image, and by Kim \etal\cite{KimECCV12W} who added relationships between objects. In more recent work, Tighe \etal\cite{TigheCVPR14} use exemplar object detectors to segment out the scene as well as individual instances. 

 There has also been work on localizing detections better using segmentation. Parkhi \etal use color models from predefined rectangles on cat and dog faces to do GrabCut and improve the predicted bounding box~\cite{ParkhiICCV11}. Dai and Hoiem generalize this to all categories and use instance and category appearance models to improve detection~\cite{DaiCVPR12}. These approaches do well when the objects are coherent in color or texture. This is not true of many categories such as people, where each object can be made of multiple regions of different appearance. An alternative to doing segmentation \emph{post facto} is to use segmentation to generate object proposals which are then classified. The proposals may be used as just bounding boxes~\cite{VanICCV11} or as region proposals~\cite{CarreiraCVPR10,ArbelaezCVPR14}. These proposals incorporate both the consistency of appearance in an object as well as the possibility of having multiple disparate regions for each object. State-of-the-art detection systems~\cite{GirshickCVPR14} and segmentation systems~\cite{CarreiraECCV12} are now based on these methods. 

In many of these approaches, segmentation is used only to localize the detections better. Other authors have explored using segmentation as a stronger cue. Fidler \etal\cite{FidlerCVPR13} use the output of a state-of-the-art semantic segmentation approach~\cite{CarreiraECCV12} to score detections better. Mottaghi~\cite{MottaghiCVPR12} uses detectors based on non-rectangular patches to both detect and segment objects.

The approaches above were typically built on features such as SIFT\cite{LoweIJCV04} or HOG\cite{DalalCVPR05}. Recently the computer vision community has shifted towards using convolutional neural networks (CNNs). CNNs have their roots in the Neocognitron proposed by Fukushima~\cite{Fukushima80}. Trained with the back-propagation algorithm, LeCun~\cite{LecunNC89} showed that they could be used for handwritten zip code recognition. They have since been used in a variety of tasks, including detection~\cite{SermanetCVPR13,SermanetICLR14} and semantic segmentation~\cite{FarabetPAMI13}. Krizhevsky \etal\cite{KrizhevskyNIPS12} showed a large increase in performance by using CNNs for classification in the ILSVRC challenge~\cite{ILSVRC12}. Donahue \etal\cite{Donahue13} showed that Krizhevsky's architecture could be used as a generic feature extractor that did well across a wide variety of tasks. Girshick \etal\cite{GirshickCVPR14} build on this and finetune Krizhevsky's architecture for detection to nearly double the state-of-the-art performance. They use a simple pipeline, using CNNs to classify bounding box proposals from~\cite{VanICCV11}. Our algorithm builds on this system, and on high quality region proposals from~\cite{ArbelaezCVPR14}.

\section{Our approach}
\label{sec:approach}
\subsection{Proposal generation}
\label{sec:prop}
A large number of methods to generate proposals have been proposed in the literature. The methods differ on the type of outputs they produce (boxes vs segments) and the metrics they do well on. Since we are interested in the AP$^r$ metric, we care about segments, and not just boxes. Keeping our task in mind, we use candidates from MCG~\cite{ArbelaezCVPR14} for this paper. This approach significantly outperforms all competing approaches on the object level Jaccard index metric, which measures the average best overlap achieved by a candidate for a ground truth object. In our experiments we find that simply switching to MCG from Selective Search~\cite{VanICCV11} improves AP$^b$ slightly (by 0.7 points), justifying this choice.

We use the proposals from MCG as is. MCG starts by computing a segmentation hierarchy at multiple image resolutions, which are then fused into a single multiscale hierarchy at the finest scale. Then candidates are produced by combinatorially grouping regions from all the single scale hierarchies and from the multiscale hierarchy. The candidates are ranked based on simple features such as size and location, shape and contour strength.

\subsection{Feature extraction}
\label{sec:feat}
We start from the R-CNN object detector proposed by Girshick \etal\cite{GirshickCVPR14} and adapt it to the SDS task. Girshick \etal  train a CNN on ImageNet Classification and then finetune the network on the PASCAL detection set. For finetuning they took bounding boxes from Selective Search, padded them, cropped them and warped them to a square and fed them to the network. Bounding boxes that overlap with the ground truth by more than 50\% were taken as positives and other boxes as negatives. The class label for each positive box was taken to be the class of the ground truth box that overlaps the most with the box. The network thus learned to predict if the bounding box overlaps highly with a ground truth bounding box. We are working with MCG instead of Selective Search, so we train a similar object detection network, finetuned using bounding boxes of MCG regions instead of Selective Search boxes. 

At test time, to extract features from a bounding box, Girshick \etal pad and crop the box, warp it to a square and pass it through the network, and extract features from one of the later layers, which is then fed into an SVM. In this paper we will use the penultimate fully connected layer.

For the SDS task, we can now use this network finetuned for detection to extract feature vectors from MCG bounding boxes. However these feature vectors do not contain any information about the actual region foreground, and so will be ill-equipped to decide if the region overlaps highly with a ground truth segmentation or not. To get around this, we start with the idea used by Girshick \etal for their experiment on semantic segmentation: we extract a second set of features from the region by feeding it the cropped, warped box, but with the background of the region masked out (with the mean image.) Concatenating these two feature vectors together gives us the feature vector we use. (In their experiments Girshick \etal found both sets of features to be useful.) This method of extracting features out of the region is the simplest way of extending the object detection system to the SDS task and forms our baseline. We call this feature extractor \textbf{\methodA{}}.

The network we are using above has been finetuned to classify bounding boxes, so its use in extracting features from the region foreground is suboptimal. Several neurons in the network may be focussing on context in the background, which will be unavailable when the network is fed the region foreground. This suggests that we should use a different network to extract the second set of features: one that is finetuned on the kinds of inputs that it is going to see. We therefore finetune another network (starting again from the net trained on ImageNet) which is fed as input cropped, padded bounding boxes of MCG regions with the background masked out. Because this region sees the actual foreground, we can actually train it to predict region overlap instead, which is what we care about. Therefore we change the labeling of the MCG regions to be based on segmentation overlap of the region with a ground truth region (instead of overlap with bounding box). We call this feature extractor \textbf{\methodB{}}.

\begin{figure}
\centering
\includegraphics[width=0.8\textwidth]{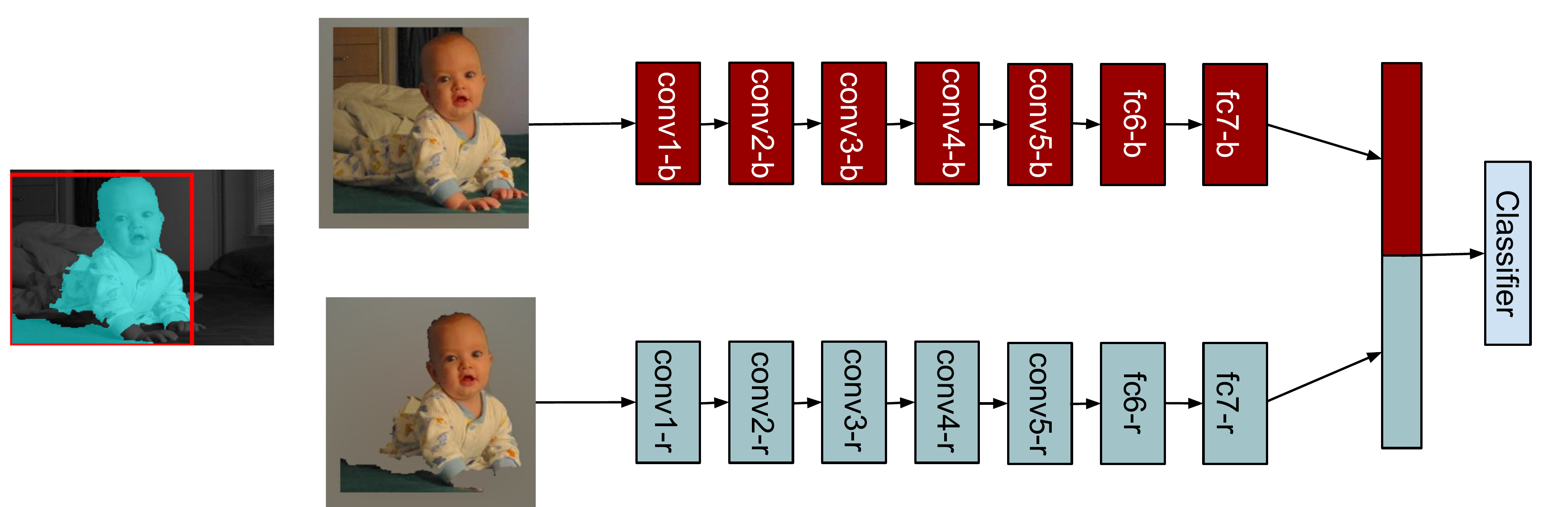}
\caption{Left: The region with its bounding box. Right: The architecture that we train for \methodC{}. The top pathway operates on cropped boxes and the bottom pathway operates on region foregrounds. }
\label{fig:net}
\end{figure}

The previous strategy is still suboptimal, because the two networks have been trained in isolation, while at test time the two feature sets are going to be combined and fed to the classifier. This suggests that one should train the networks jointly. We formalize this intuition as follows. We create a neural network with the architecture shown in Figure~\ref{fig:net}. This architecture is a single network with two pathways. The first pathway operates on the cropped bounding box of the region (the ``box" pathway) while the second pathway operates on the cropped bounding box with the background masked (the ``region" pathway). The two pathways are disjoint except at the very final classifier layer, which concatenates the features from both pathways. Both these pathways individually have the same architecture as that of Krizhevsky \etal Note that both \methodA{} and \methodB{} can be seen as instantiations of this architecture, but with different sets of weights. \methodA{} uses the same network parameters for both pathways. For \methodB{}, the box pathway gets its weights from a network finetuned separately using bounding box overlap, while the region pathway gets its parameters from a network finetuned separately using region overlap. 

Instead of using the same network in both pathways or training the two pathways in isolation, we now propose to train it as a whole directly. We use segmentation overlap as above. We initialize the box pathway with the network finetuned on boxes and the region pathway with the network finetuned on regions, and then finetune the entire network. At test time, we discard the final classification layer and use the output of the penultimate layer, which concatenates the features from the two pathways. We call this feature extractor \textbf{\methodC{}}.

\subsection{Region classification}
\label{sec:classify}
We use the features from the previous step to train a linear SVM. We first train an initial SVM using ground truth as positives and regions overlapping ground truth by less than 20\% as negative. Then we re-estimate the positive set: for each ground truth we pick the highest scoring MCG candidate that overlaps by more than 50\%. Ground truth regions for which no such candidate exists (very few in number) are discarded. We then retrain the classifier using this new positive set. This training procedure corresponds to a multiple instance learning problem where each ground truth defines a positive bag of regions that overlap with it by more than 50\%, and each negative region is its own bag. We found this training to work better than using just the ground truth as positives.

At test time we use the region classifiers to score each region. Because there may be multiple overlapping regions, we do a strict non-max suppression using a region overlap threshold of 0. This is because while the bounding box of two objects can in fact overlap, their pixel support in the image typically shouldn't. Post NMS, we work with only the top 20,000 detections for each category (over the whole dataset) and discard the rest for computational reasons. We confirmed that this reduction in detections has no effect on the AP$^r$ metric. 
\subsection{Region refinement}
\label{sec:refine}

\begin{figure}
\includegraphics[width=0.15\textwidth]{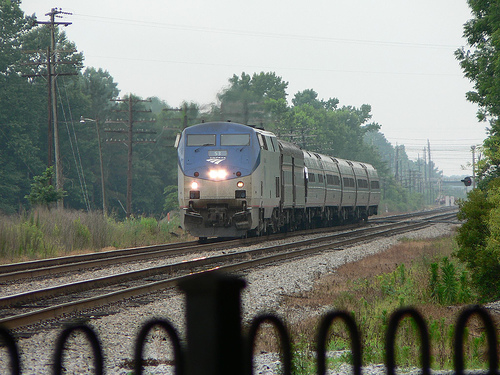}
\includegraphics[width=0.15\textwidth]{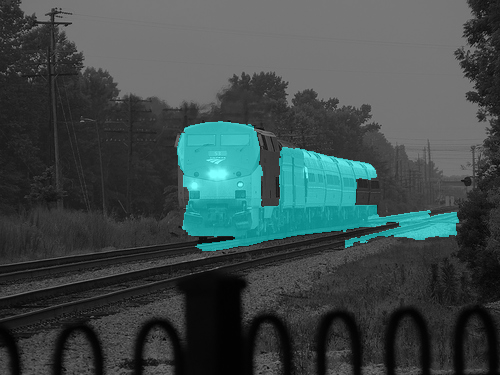}
\includegraphics[width=0.15\textwidth]{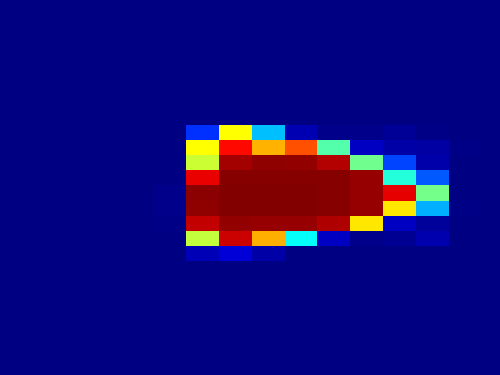}
\includegraphics[width=0.15\textwidth]{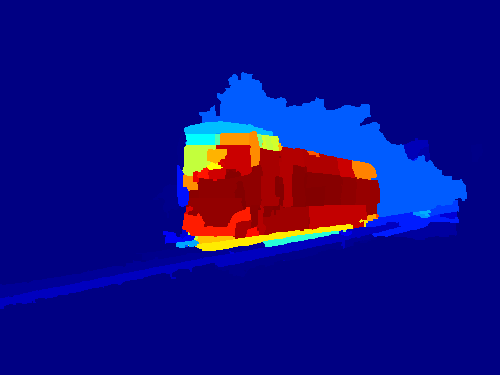}
\includegraphics[width=0.15\textwidth]{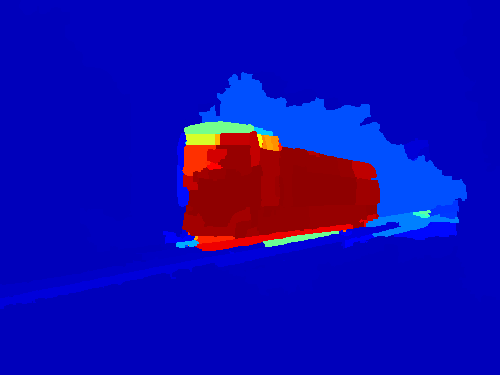}
\includegraphics[width=0.15\textwidth]{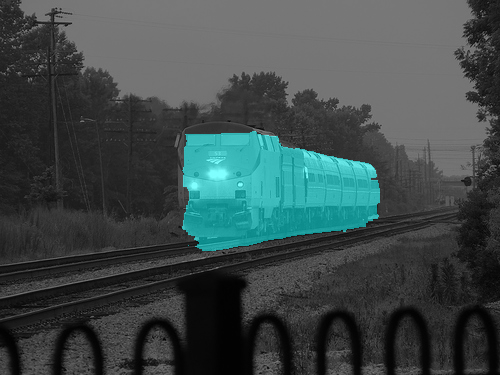}\\
\includegraphics[width=0.15\textwidth]{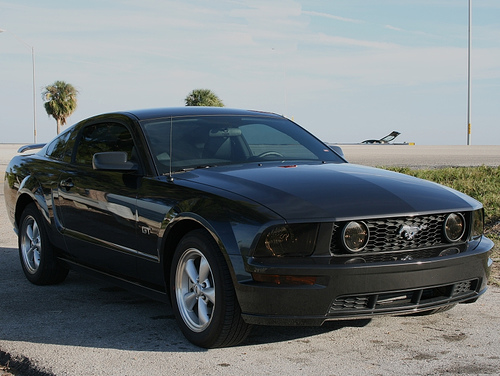}
\includegraphics[width=0.15\textwidth]{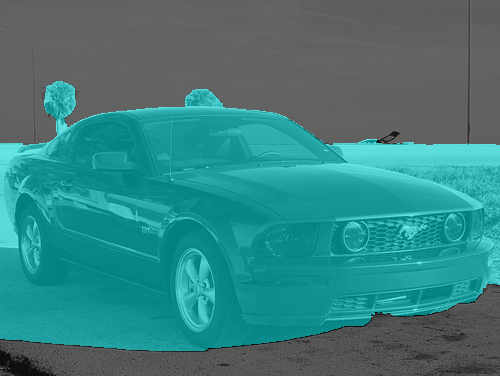}
\includegraphics[width=0.15\textwidth]{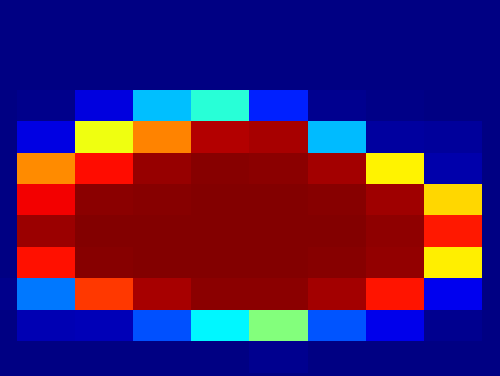}
\includegraphics[width=0.15\textwidth]{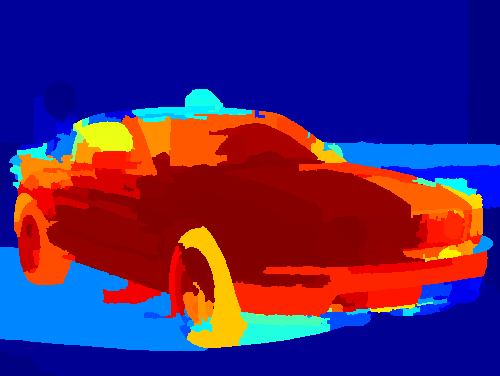}
\includegraphics[width=0.15\textwidth]{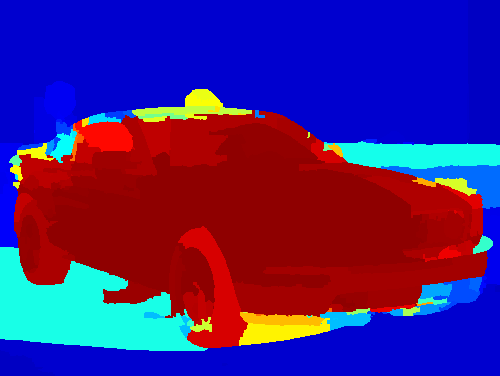}
\includegraphics[width=0.15\textwidth]{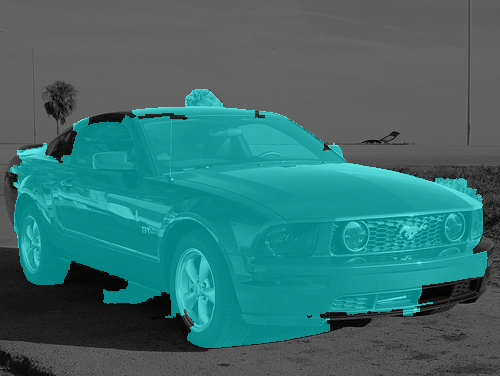}\\
\includegraphics[width=0.15\textwidth]{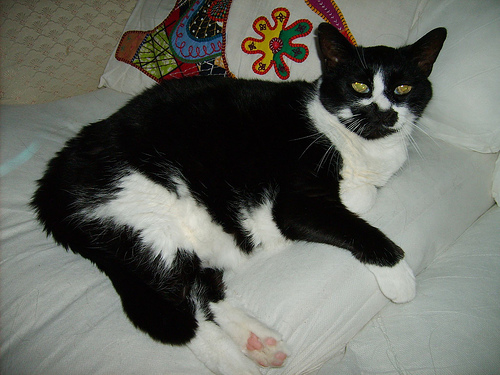}
\includegraphics[width=0.15\textwidth]{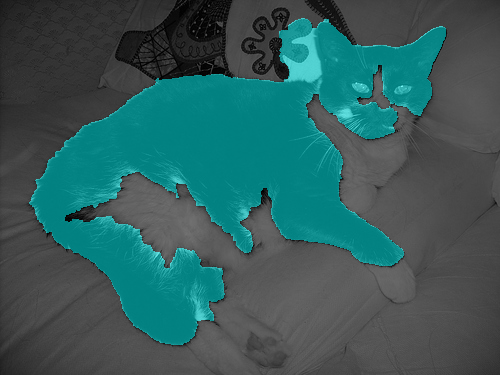}
\includegraphics[width=0.15\textwidth]{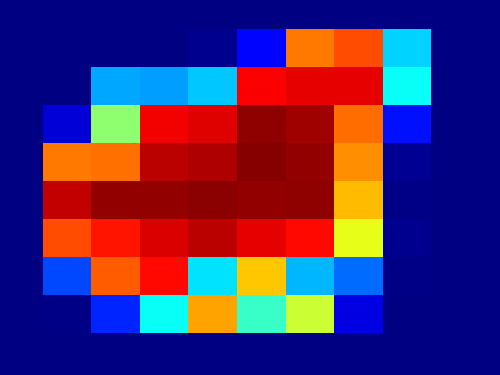}
\includegraphics[width=0.15\textwidth]{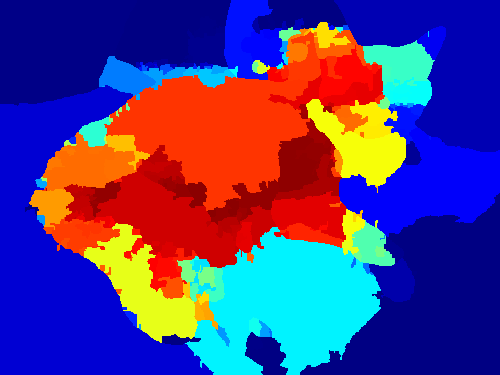}
\includegraphics[width=0.15\textwidth]{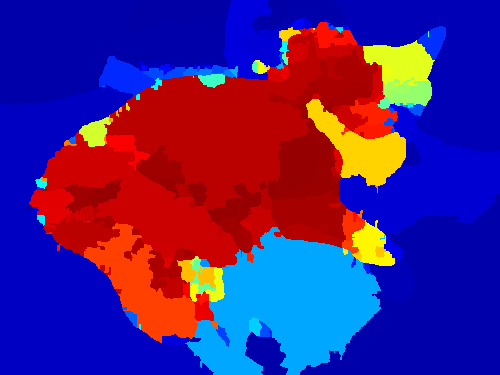}
\includegraphics[width=0.15\textwidth]{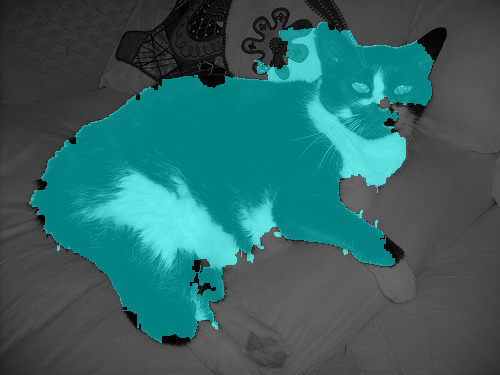}
\caption{Some examples of region refinement. We show in order the image, the original region, the coarse $10 \times 10$ mask, the coarse mask projected to superpixels, the output of the final classifier on superpixels and the final region after thresholding. Refinement uses top-down category specific information to fill in the body of the train and the cat and remove the road from the car.}
\label{fig:refinement}
\end{figure}
We take each of the remaining regions and refine its support. This is necessary because our region candidates have been created by a purely bottom-up, class agnostic process. Since the candidate generation has not made use of category-specific shape information, it is prone to both undershooting (\ie missing some part of the object) and overshooting (\ie including  extraneous stuff).

We first learn to predict a coarse, top-down figure-ground mask for each region. To do this, we take the bounding box of each predicted region, pad it as for feature extraction, and then discretize the resulting box into a $10\times 10$ grid. For each grid cell we train a logistic regression classifier to predict the probability that the grid cell belongs to the foreground. The features we use are the features extracted from the CNN, together with the figure-ground mask of the region discretized to the same $10 \times 10$ grid. The classifiers are trained on regions from the training set that overlap by more than 70\% with a ground truth region. 

This coarse figure-ground mask makes a top-down prediction about the shape of the object but does not necessarily respect the bottom-up contours. In addition, because of its coarse nature it cannot do a good job of modeling thin structures like aircraft wings or structures that move around. This information needs to come from the bottom-up region candidate. Hence we train a second stage to combine this coarse mask with the region candidate. We project the coarse mask to superpixels by assigning to each superpixel the average value of the coarse mask in the superpixel. Then we classify each superpixel, using as features this projected value in the superpixel and a 0 or 1 encoding if the superpixel belongs to the original region candidate.  
Figure~\ref{fig:refinement} illustrates this refinement.
  
\section{Experiments and results}
\label{sec:experiments}
We use the segmentation annotations from SBD~\cite{HariharanICCV11} to train and evaluate. We train all systems on PASCAL VOC 2012 train. For all training and finetuning of the network we use the recently released Caffe framework~\cite{Jia13}.  

\subsection{Results on AP$^r$ and AP$^r_{vol}$}

Table~\ref{table:APr} and Table~\ref{table:APrvol} show results on the AP$^r$ and the AP$^r_{vol}$ metrics respectively on PASCAL VOC 2012 val (ground truth segmentations are not available for test). We compute AP$^r_{vol}$ by averaging the AP$^r$ obtained for 9 thresholds. 
\begin{enumerate}
\item \textbf{O$_2$P} uses features and regions from Carreira \etal\cite{CarreiraECCV12}, which is the state-of-the-art in semantic segmentation. We train region classifiers on these features and do NMS to get detections. This baseline gets a mean AP$^r$ of 25.2\% and a mean AP$^r_{vol}$ of 23.4\%. 
\item \textbf{\methodA{}} is our most naive feature extractor. It uses MCG candidates and features from the bounding box and region foreground, using a single CNN finetuned using box overlaps. It achieves a mean AP$^r$ of 42.9\% and a mean AP$^r_{vol}$ of 37.0\%, a large jump over O$_2$P. This mirrors gains in object detection observed by Girshick \etal\cite{GirshickCVPR14}, although since O$_2$P is not designed for this task the comparison is somewhat unfair. 
\item \textbf{\methodB{}} is the result of finetuning a separate network exclusively on region foregrounds with labels defined by region overlap. This gives a large jump of the AP$^r$ metric (of about 4 percentage points) and a smaller but significant jump on the AP$^r_{vol}$ metric of about 2.5 percentage points. 
\item \textbf{\methodC{}} is the result of training a single large network with two pathways. There is a clear gain over using two isolated networks: on both metrics we gain about 0.7 percentage points.
\item \textbf{\methodC{}+ref} is the result of refining the masks of the regions obtained from \methodC{}. We again gain 2 points in the AP$^r$ metric and 1.2 percentage points in the AP$^r_{vol}$ metric. This large jump indicates that while MCG candidates we start from are very high quality, there is still a lot to be gained from refining the regions in a category specific manner.
\end{enumerate}
A paired sample t-test indicates that each of the above improvements are statistically significant at the 0.05 significance level.

\begin{figure}
\centering
\includegraphics[width=0.45\textwidth]{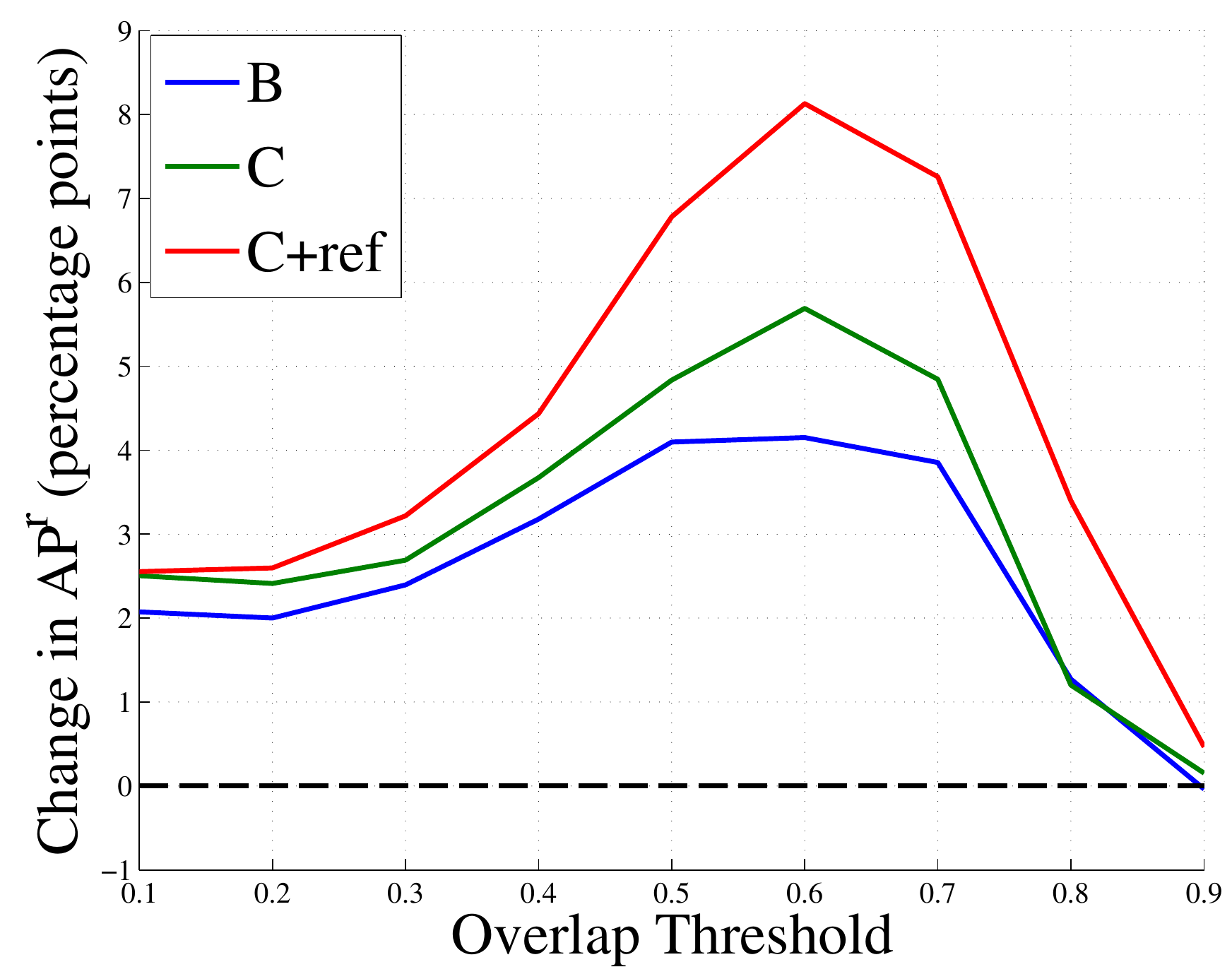}
\includegraphics[width=0.45\textwidth]{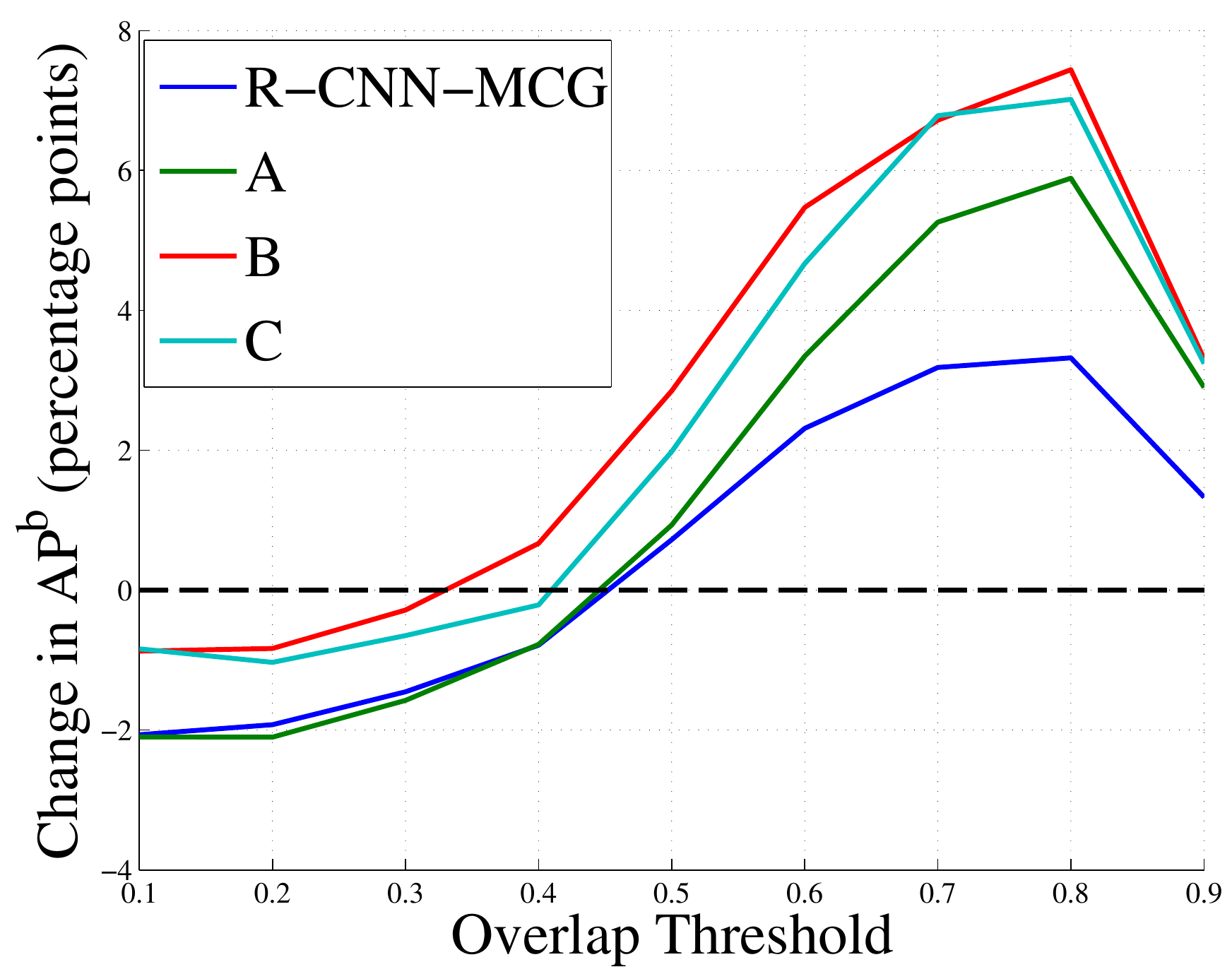}
\caption{Left: Improvement in mean AP$^r$ over \methodA{} due to our 3 variants for a variety of overlap thresholds. We get improvements for all overlap thresholds. Right: A similar plot for AP$^b$. Improvements are relative to R-CNN with Selective Search proposals~\cite{GirshickCVPR14}. As the threshold becomes stricter, the better localization of our approach is apparent.}
\label{improvovthresh}
\end{figure}

The left part of Figure~\ref{improvovthresh} plots the improvement in mean AP$^r$ over \methodA{}  as we vary the threshold at which a detection is considered correct. Each of our improvements increases AP$^r$ across all thresholds, indicating that we haven't overfit to a particular regime.

Clearly we get significant gains over both our naive baseline as well as O2P. However, prior approaches that reason about segmentation together with detection might do better on the AP$^r$ metric. To see if this is the case, we compare to the SegDPM work of Fidler \etal\cite{FidlerCVPR13}. SegDPM combined DPMs~\cite{FelzenszwalbPAMI10} with O$_2$P~\cite{CarreiraECCV12} and achieved a 9 point boost over DPMs in classical object detection. For this method, only the bounding boxes are available publicly, and for some boxes the algorithm may choose not to have associated segments. We therefore compute an upper bound of its performance by taking each detection, considering all MCG regions whose bounding box overlaps with the detection by more than 70\%, and selecting the region which best overlaps a ground truth. 

Since SegDPM detections are only available on PASCAL VOC2010 val, we restrict our evaluations only to this set.  Our upper bound on SegDPM has a mean AP$^r$ of \textbf{31.3}, whereas \methodC{}+ref achieves a mean AP$^r$ of \textbf{50.3}.
\begin{table}
\centering
\caption{Results on AP$^r$ on VOC2012 val. All numbers are \%.}
\small{
\renewcommand{\tabcolsep}{1.5mm}
\begin{tabular}{l|ccccc}
 & O$_2$P & \methodA{} & \methodB{} & \methodC{} & \methodC{}+ref\\
\hline
 aeroplane & 56.5 & 61.8 & 65.7 & 67.4 & \textbf{68.4}\\
 bicycle & 19.0 & 43.4 & \textbf{49.6} & \textbf{49.6} & 49.4\\
 bird & 23.0 & 46.6 & 47.2 & 49.1 & \textbf{52.1}\\
 boat & 12.2 & 27.2 & 30.0 & 29.9 & \textbf{32.8}\\
 bottle & 11.0 & 28.9 & 31.7 & 32.0 & \textbf{33.0}\\
 bus & 48.8 & 61.7 & 66.9 & 65.9 & \textbf{67.8}\\
 car & 26.0 & 46.9 & 50.9 & 51.4 & \textbf{53.6}\\
 cat & 43.3 & 58.4 & 69.2 & 70.6 & \textbf{73.9}\\
 chair & 4.7 & 17.8 & 19.6 & \textbf{20.2} & 19.9\\
 cow & 15.6 & 38.8 & 42.7 & 42.7 & \textbf{43.7}\\
 diningtable & 7.8 & 18.6 & 22.8 & 22.9 & \textbf{25.7}\\
 dog & 24.2 & 52.6 & 56.2 & 58.7 & \textbf{60.6}\\
 horse & 27.5 & 44.3 & 51.9 & 54.4 & \textbf{55.9}\\
 motorbike & 32.3 & 50.2 & 52.6 & 53.5 & \textbf{58.9}\\
 person & 23.5 & 48.2 & 52.6 & 54.4 & \textbf{56.7}\\
 pottedplant & 4.6 & 23.8 & 25.7 & 24.9 & \textbf{28.5}\\
 sheep & 32.3 & 54.2 & 54.2 & 54.1 & \textbf{55.6}\\
 sofa & 20.7 & 26.0 & \textbf{32.2} & 31.4 & 32.1\\
 train & 38.8 & 53.2 & 59.2 & 62.2 & \textbf{64.7}\\
 tvmonitor & 32.3 & 55.3 & 58.7 & 59.3 & \textbf{60.0}\\
\hline
 Mean & 25.2 & 42.9 & 47.0 & 47.7 & \textbf{49.7}\\
\end{tabular}
}
\label{table:APr}
\end{table}
\begin{table}
\centering
\caption{Results on AP$^r_{vol}$ on VOC2012 val. All numbers are \%.}
\small{
\renewcommand{\tabcolsep}{1.5mm}
\begin{tabular}{l|ccccc}
 & O$_2$P & \methodA{} & \methodB{} & \methodC{} & \methodC{}+ref\\
\hline
 aeroplane & 46.8 & 48.3 & 51.1 & \textbf{53.2} & 52.3\\
 bicycle & 21.2 & 39.8 & 42.1 & 42.1 & \textbf{42.6}\\
 bird & 22.1 & 39.2 & 40.8 & 42.1 & \textbf{42.2}\\
 boat & 13.0 & 25.1 & 27.5 & 27.1 & \textbf{28.6}\\
 bottle & 10.1 & 26.0 & 26.8 & 27.6 & \textbf{28.6}\\
 bus & 41.9 & 49.5 & 53.4 & 53.3 & \textbf{58.0}\\
 car & 24.0 & 39.5 & 42.6 & 42.7 & \textbf{45.4}\\
 cat & 39.2 & 50.7 & 56.3 & 57.3 & \textbf{58.9}\\
 chair & 6.7 & 17.6 & 18.5 & 19.3 & \textbf{19.7}\\
 cow & 14.6 & 32.5 & 36.0 & 36.3 & \textbf{37.1}\\
 diningtable & 9.9 & 18.5 & 20.6 & 21.4 & \textbf{22.8}\\
 dog & 24.0 & 46.8 & 48.9 & 49.0 & \textbf{49.5}\\
 horse & 24.4 & 37.7 & 41.9 & \textbf{43.6} & 42.9\\
 motorbike & 28.6 & 41.1 & 43.2 & 43.5 & \textbf{45.9}\\
 person & 25.6 & 43.2 & 45.8 & 47.0 & \textbf{48.5}\\
 pottedplant & 7.0 & 23.4 & 24.8 & 24.4 & \textbf{25.5}\\
 sheep & 29.0 & 43.0 & 44.2 & 44.0 & \textbf{44.5}\\
 sofa & 18.8 & 26.2 & 29.7 & 29.9 & \textbf{30.2}\\
 train & 34.6 & 45.1 & 48.9 & 49.9 & \textbf{52.6}\\
 tvmonitor & 25.9 & 47.7 & 48.8 & 49.4 & \textbf{51.4}\\
\hline
 Mean & 23.4 & 37.0 & 39.6 & 40.2 & \textbf{41.4}\\
\end{tabular}
}
\label{table:APrvol}
\end{table}
\subsection{Producing diagnostic information}
Inspired by~\cite{HoiemECCV12}, we created tools for figuring out error modes and avenues for improvement for the SDS task. As in~\cite{HoiemECCV12}, we evaluate the impact of error modes by measuring the improvement in AP$^r$ if the error mode was corrected. For localization, we assign labels to detections under two thresholds: the usual strict threshold of 0.5 and a more lenient threshold of 0.1 (note that this is a threshold on region overlap). Detections that count as true positives under the lenient threshold but as false positives under the strict threshold are considered mislocalizations. Duplicate detections are also considered mislocalizations. We then consider the performance if either a) all mislocalized instances were removed, or b) all mislocalized instances were correctly localized and duplicates removed. 

\begin{figure}
\centering
\includegraphics[width=0.45\textwidth]{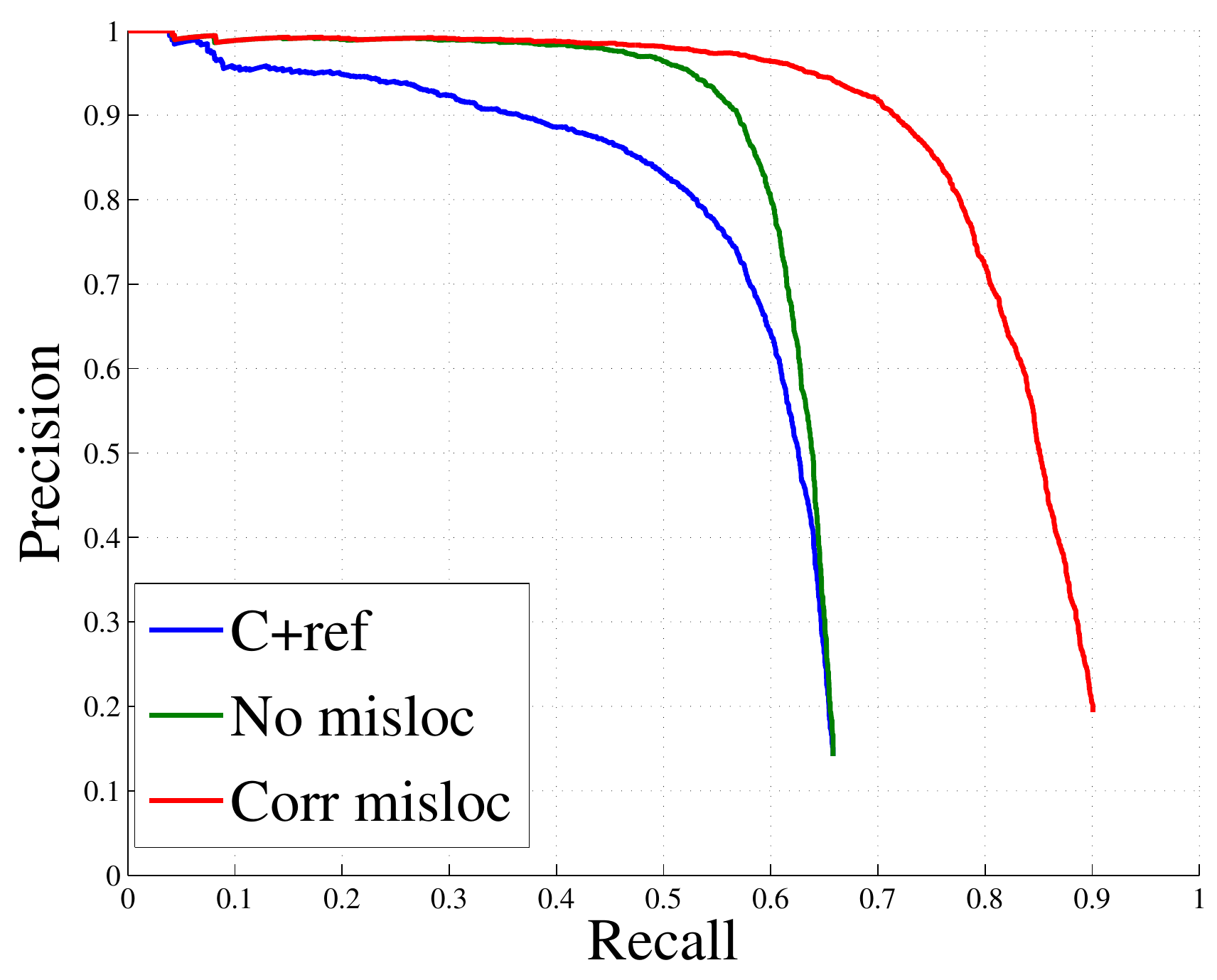}
\includegraphics[width=0.45\textwidth]{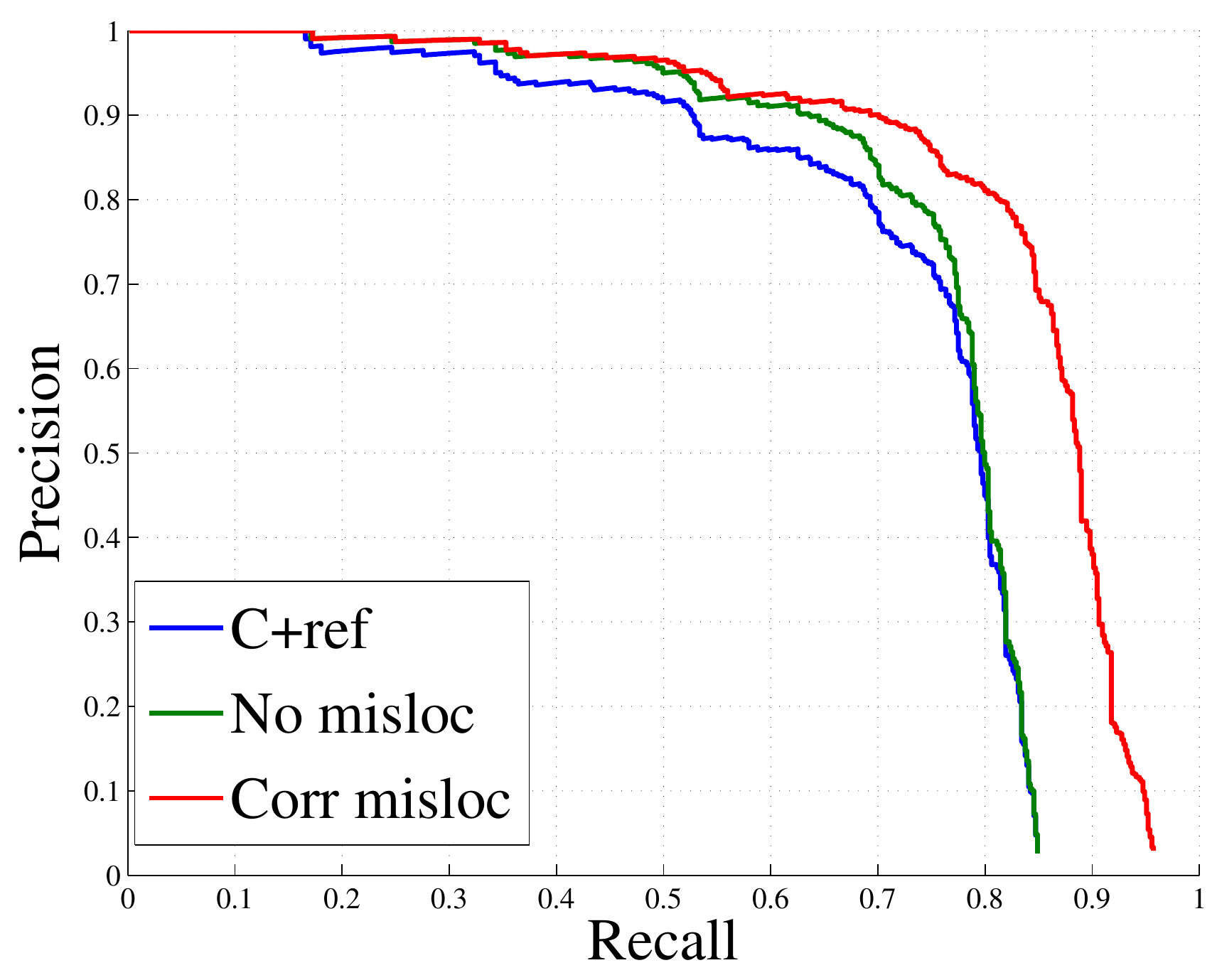}
\caption{PR on person(left) and cat(right). Blue is \methodC{}+ref. Green is if an oracle removes mislocalized predictions, and red is if the oracle corrects our mislocalizations.}
\label{personPR}
\end{figure}

Figure~\ref{personPR} shows how the PR curve for the AP$^r$ benchmark changes if mislocalizations are corrected or removed for two categories. For the person category, removing mislocalizations brings precision up to essentially 100\%, indicating that mislocalization is the predominant source of false positives. Correcting the mislocalizations provides a huge jump in recall. For the cat category the improvement provided by better localization is much less, indicating that there are still some false positives arising from misclassifications.

\begin{figure}
\centering
\includegraphics[width=0.43\textwidth]{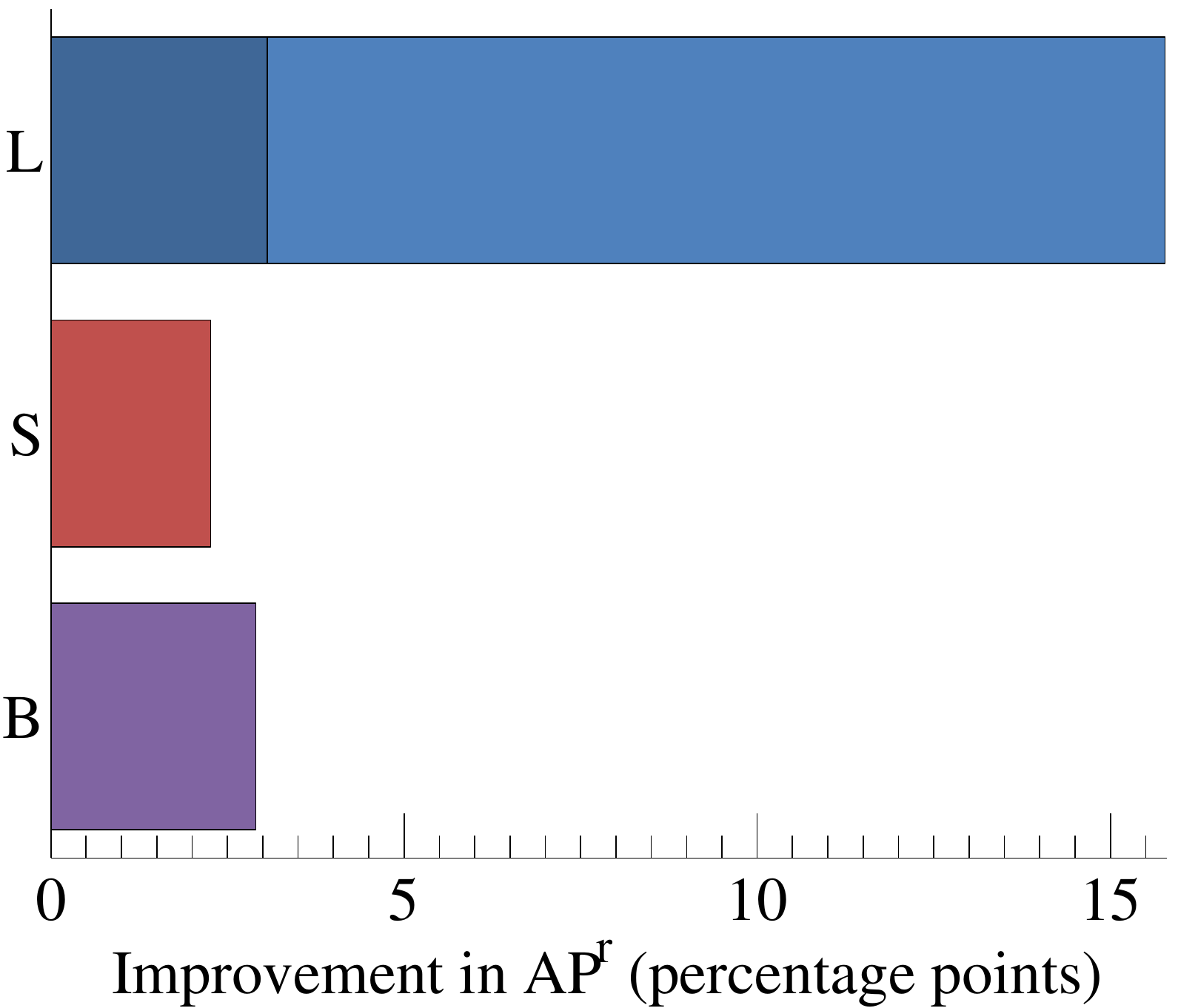}
\includegraphics[width=0.46\textwidth]{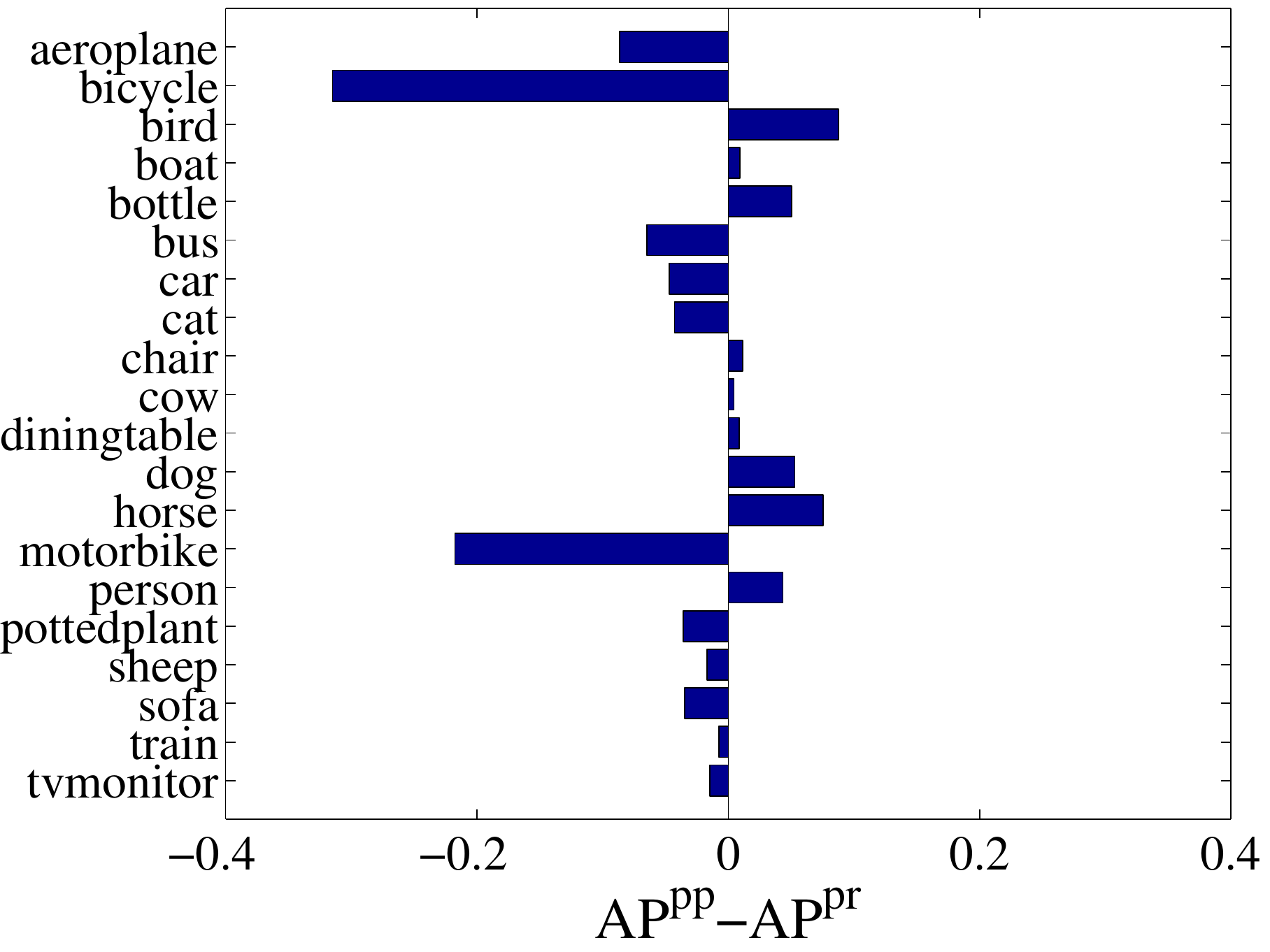}
\caption{Left: Impact of the three kinds of false positives on mean AP$^r$.  L : mislocalization, B : detection on background, and S : misfirings on similar categories. Right: Disambiguating between two kinds of mislocalizations. Bars to the left mean that we frequently overshoot the ground truth, while bars to the right mean that we undershoot. .}
\label{misloc}
\end{figure}

We can do this analysis for all categories. The average improvement in AP$^r$ by fixing mislocalization is a measure of the impact of mislocalization on performance. We can also measure impact in this way for other error modes: for instance, false positives on objects of other similar categories, or on background~\cite{HoiemECCV12}. (For defining similar and non-similar categories, we divide object categories into ``animals", ``transport" and ``indoor" groups.) The left subfigure in Figure~\ref{misloc} shows the result of such an analysis on our best system (\methodC{}+ref). The dark blue bar shows the AP$^r$ improvement if we remove mislocalized detections and the light blue bar shows the improvement if we correct them. The other two bars show the improvement from removing confusion with similar categories and background. Mislocalization has a huge impact: it sets us back by about 16 percentage points. Compared to that confusion with similar categories or background is virtually non-existent.

We can measure the impact of mislocalization on the other algorithms in Table~\ref{table:APr} as well, as shown in Table~\ref{table:loc}. It also shows the upper bound AP$^r$ achievable when all mislocalization is fixed. Improvements in the feature extractor improve the upper bound (indicating fewer misclassifications) but also reduce the gap due to mislocalization (indicating better localization). Refinement doesn't change the upper bound and only improves localization, as expected.

To get a better handle on what one needs to do to improve localization, we considered two statistics. For each detection and a ground truth, instead of just taking the overlap (\ie intersection over union), we can compute the pixel precision (fraction of the region that lies inside the ground truth) and pixel recall (fraction of the ground truth that lies inside the region). It can be shown that having both a pixel precision $>$ 67\% and a pixel recall $>$ 67\% is guaranteed to give an overlap of greater than 50\%. We assign detection labels using pixel precision or pixel recall using a threshold of 67\% and compute the respective AP. Comparing these two numbers then gives us a window into the kind of localization errors: a low pixel precision AP indicates that the error mode is overshooting the region and predicting extraneous background pixels, while a low pixel recall AP indicates that the error mode is undershooting the region and missing out some ground truth pixels. 

The second half of Figure~\ref{misloc} shows the difference between pixel precision AP (AP$^{pp}$) and pixel recall AP (AP$^{pr}$). Bars to the left indicate higher pixel recall AP, while bars to the right indicate higher pixel precision AP. For some categories such as person and bird we tend to miss ground truth pixels, whereas for others such as bicycle we tend to leak into the background.


\begin{table}
\centering
\caption{Maximum achievable AP$^r$ (assuming perfect localization) and loss in AP$^r$ due to mislocalization for all systems.}
\renewcommand{\tabcolsep}{1.5mm}
\begin{tabular}{l|cccc}
& \methodA{} & \methodB{} & \methodC{} & \methodC{}+ref\\
\hline
AP Upper bound & 63.0 & 65.0 & 65.4 & 65.5\\
Loss due to mislocalization & 20.1 & 18.0 & 17.7 & 15.8
\end{tabular}
\label{table:loc}
\end{table}
%\vspace{-0.1in}
\subsection{Results on AP$^b$ and AP$^b_{vol}$}
Comparison with prior work is easier on the classical bounding box and segmentation metrics. It also helps us evaluate if handling the SDS task also improves performance on the individual tasks.
To compare on AP$^b$, we retrain our final region classifiers for the bounding box detection task. This is because the ranking of regions based on bounding box overlap is different from that based on segmentation overlap. As in~\cite{GirshickCVPR14}, we use ground truth boxes as positive, and MCG boxes overlapping by less than 50\% as negative. At test time we do not do any region refinement.

We add two baselines: R-CNN is the system of Girshick \etal taken as is, and R-CNN-MCG is R-CNN on boxes from MCG instead of Selective Search. Note that neither of these baselines uses features from the region foreground.

Table~\ref{APb} shows the mean AP$^b$ and AP$^b_{vol}$. We get improvements over R-CNN on both AP$^b$ and AP$^b_{vol}$, with improvements on the latter metric being somewhat larger. The right half of Figure~\ref{improvovthresh} shows the variation in AP$^b$ as we vary the overlap threshold for counting something as correct. We plot the improvement in AP$^b$ over vanilla R-CNN. We do worse than R-CNN for low thresholds, but are much better for higher thresholds. This is also true to some extent for R-CNN-MCG, so this is partly a property of MCG, and partly a consequence of our algorithm's improved localization. Interestingly, \methodC{} does worse than \methodB{}. We posit that this is because now the entire network has been finetuned for SDS.

Finally we evaluated \methodC{} on PASCAL VOC 2012 test. Our mean AP$^b$ of \textbf{50.7} is an improvement over the R-CNN mean AP$^b$ of \textbf{49.6} (both without bounding box regression), and much better than other systems, such as  SegDPM~\cite{FidlerCVPR13} (\textbf{40.7}). 
%\vspace{-0.1in}
\begin{table}
\centering
\caption{ Results on AP$^b$ and AP$^b_{vol}$ on VOC12 val. All numbers are \%.}
\begin{tabular}{l|ccccc}
& R-CNN\cite{GirshickCVPR14} & R-CNN-MCG & \methodA{} &\methodB{}& \methodC{}\\
\hline
Mean AP$^b$ & 51.0 & 51.7 & 51.9 \; & \textbf{53.9} \; & 53.0 \\
Mean AP$^b_{vol}$& 41.9 & 42.4 & 43.2 \; & \textbf{44.6} \; & 44.2
\end{tabular}
\label{APb}
\end{table}
%\vspace{-0.3in}
\subsection{Results on pixel IU}
For the semantic segmentation task, we convert the output of our final system (\methodC{}+ref) into a pixel-level category labeling using the simple pasting scheme proposed by Carreira \etal\cite{CarreiraECCV12}. We cross validate the hyperparameters of this pasting step on the VOC11 segmentation Val set. The results are  in Table~\ref{pixelIU}. We compare to O$_2$P~\cite{CarreiraECCV12} and R-CNN which are the current state-of-the-art on this task. We advance the state-of-the-art by about 5 points, or 10\% relative.

\begin{figure}
\centering
\includegraphics[width=0.085\textwidth]{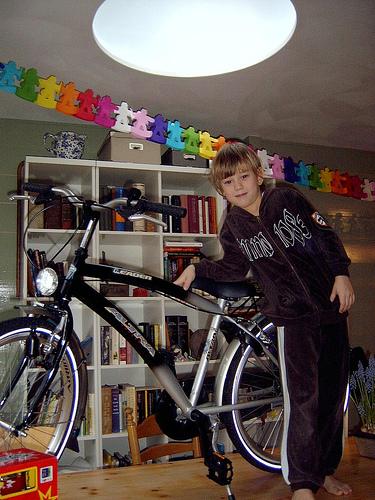}
\includegraphics[width=0.085\textwidth]{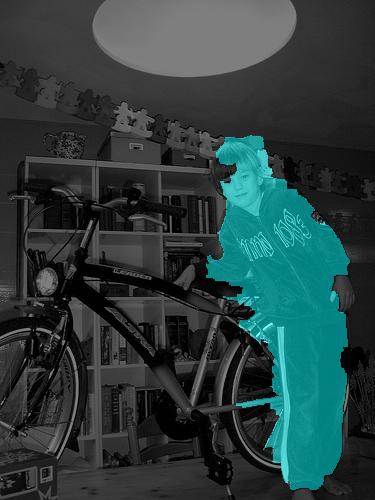}
\includegraphics[width=0.14\textwidth]{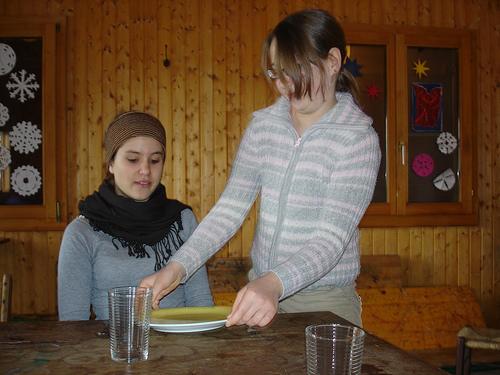}
\includegraphics[width=0.14\textwidth]{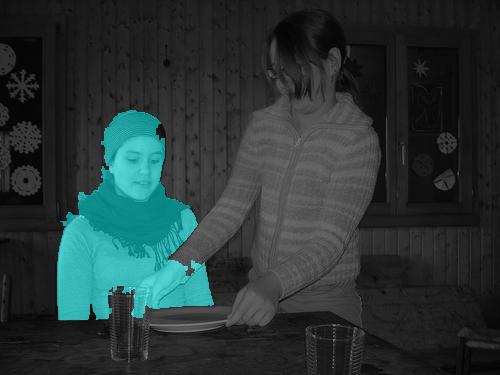}
\includegraphics[width=0.085\textwidth]{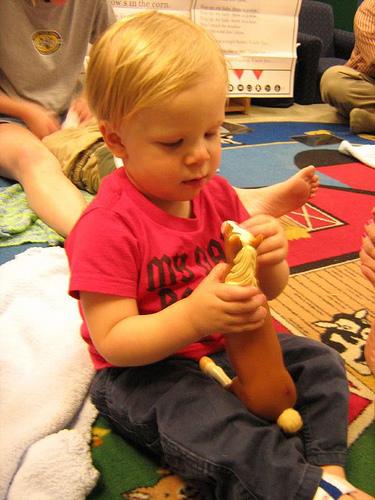}
\includegraphics[width=0.085\textwidth]{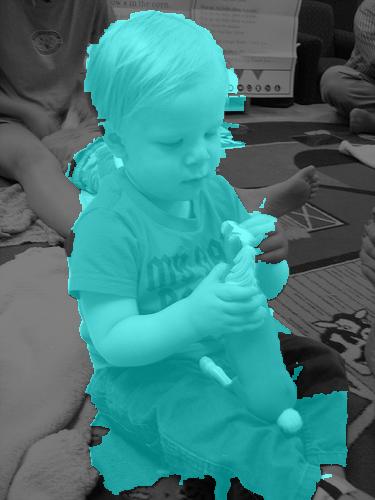}\\
\includegraphics[width=0.14\textwidth]{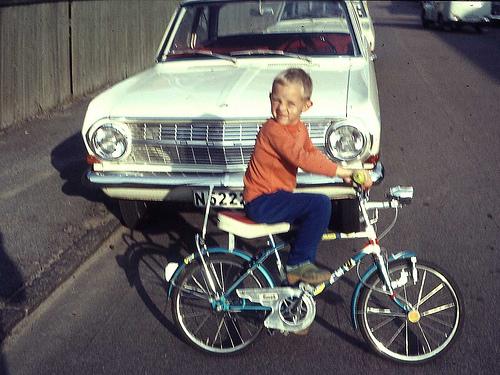}
\includegraphics[width=0.14\textwidth]{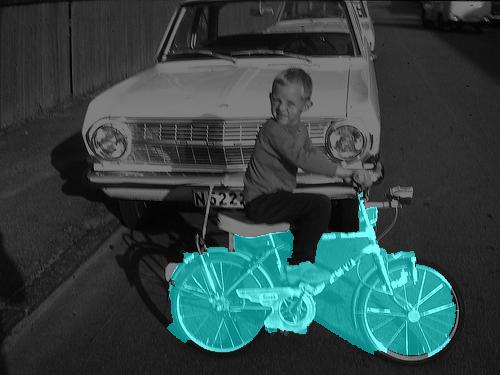}
\includegraphics[width=0.14\textwidth]{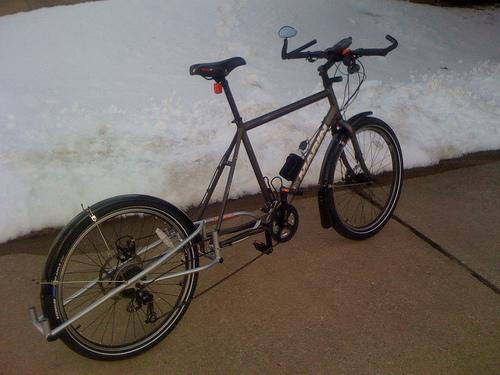}
\includegraphics[width=0.14\textwidth]{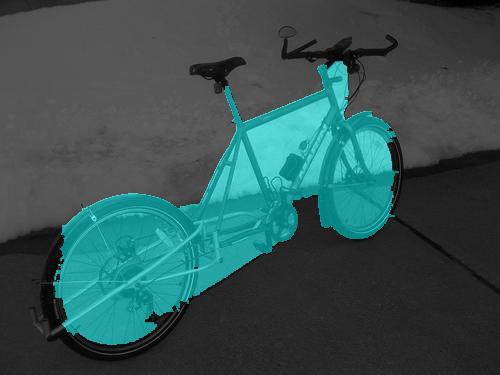}
\includegraphics[width=0.14\textwidth]{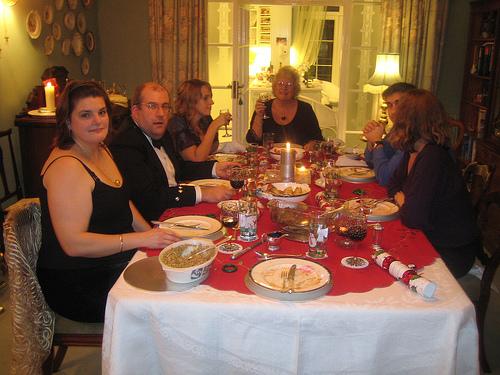}
\includegraphics[width=0.14\textwidth]{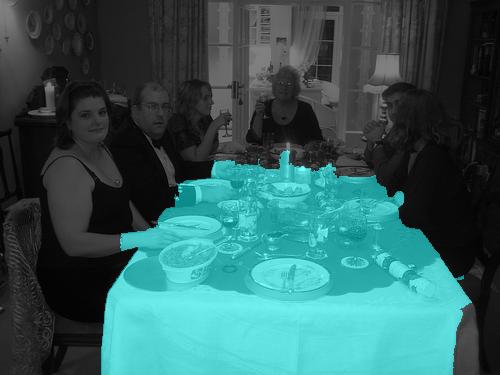}\\
\includegraphics[width=0.14\textwidth]{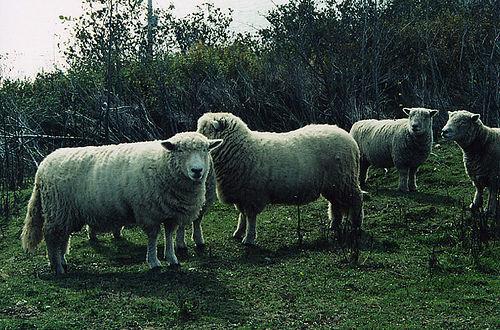}
\includegraphics[width=0.14\textwidth]{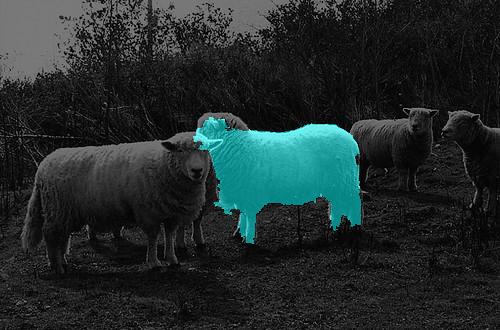}
\includegraphics[width=0.14\textwidth]{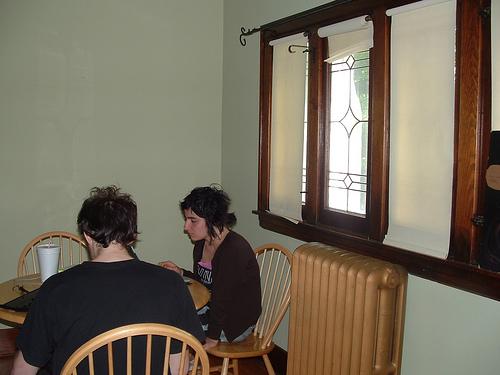}
\includegraphics[width=0.14\textwidth]{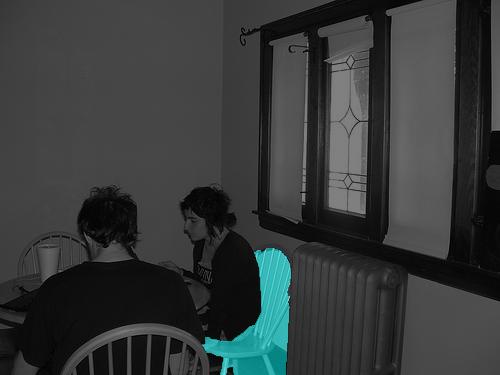}
\includegraphics[width=0.14\textwidth]{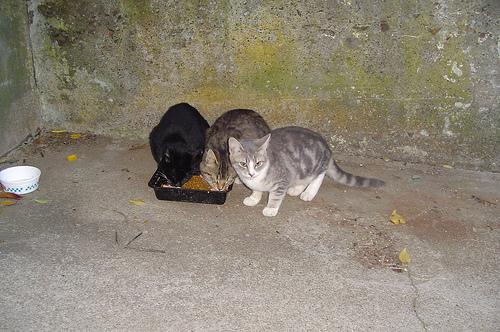}
\includegraphics[width=0.14\textwidth]{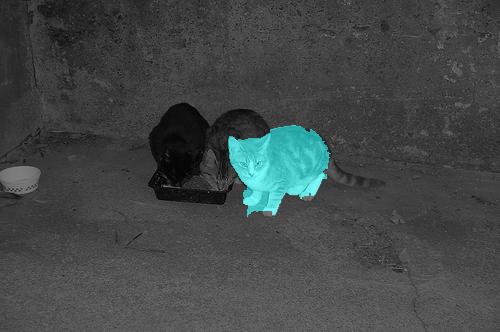}
%\vspace{-0.1in}
\caption{  Top detections:  3 persons, 2 bikes, diningtable, sheep, chair, cat. We can handle uncommon pose and clutter and are able to resolve individual instances. }
\label{fig:examples}
\end{figure}

To conclude, our pipeline achieves good results on the SDS task while improving state-of-the-art in object detection and semantic segmentation. Figure~\ref{fig:examples} shows examples of the output of our system.
%\vspace{-0.3in}
\begin{table}
\centering
\caption{Results on Pixel IU. All numbers are \%.}
\small{
\begin{tabular}{l|cccc}
& O$_2$P~\cite{CarreiraECCV12} & R-CNN~\cite{GirshickCVPR14} & \methodC{}+ref\\
\hline
Mean Pixel IU (VOC2011 Test) & 47.6 & 47.9 & \textbf{52.6} \\
Mean Pixel IU (VOC2012 Test) & 47.8 & - & \textbf{51.6}\\
\end{tabular}
}
\label{pixelIU}
\end{table}

%\vspace{-0.8in}
\scriptsize{
\noindent\textbf{Acknowledgments.}
This work was supported by ONR MURI N000141010933, a Google Research Grant and a Microsoft Research fellowship. We thank the NVIDIA Corporation for providing GPUs through their academic program.
}
 \clearpage

\bibliographystyle{splncs03}
\bibliography{main}
\end{document}